# Contour Sparse Representation with SDD Features for Object Recognition

ZhenZhou Wang

*Abstract*—Slope difference distribution (SDD) is computed for the one-dimensional curve. It is not only robust to calculate the partitioning point to separate the curve logically, but also robust to calculate the clustering center of each part of the separated curve. SDD has been proposed for image segmentation and it outperforms all existing image segmentation methods. For verification purpose, we have made the Matlab codes of comparing SDD method with existing image segmentation methods freely available at Matlab Central. The contour of the object is similar to the histogram of the image. Thus, feature detection by SDD from the contour of the object is also feasible. In this letter, SDD features are defined and they form the sparse representation of the object's contour. The reference model of each object is built based on the SDD features and then model matching is used for on line object recognition. The experimental results are very encouraging. For the gesture recognition, SDD achieved 100% accuracy for two public datasets: the NUS dataset and the near-infrared dataset. For the object recognition, SDD achieved 100% accuracy for the Kimia 99 dataset.

## I. Introduction

SLOPE difference distribution (SDD) has been proposed for image segmentation by calculating the clustering centers of different pixel classes and the partitioning points between adjacent pixel classes [1-3]. Specifically, SDD based image segmentation method has achieved significantly better accuracies than existing image segmentation methods in segmenting the left ventricles in magnetic resonance images [4-8]. Interested readers could find the MATLAB source codes at [6, 8] for verification. In this letter, SDD is applied to detect the features for object recognition that is one of the most challenging tasks in computer vision [9-12]. The traditional object recognition methods could be divided into two categories: One is the moment based method [9-10] and the other is the Fourier descriptor based method [11-12]. However, the recognition accuracies of these two categories of methods are not satisfactory as reported in [13].

Since human beings can recognize the objects easily solely based on the contour information of the objects, there must be a robust method that could recognize the object solely based on the contour of the object. However, the whole contour usually contains too much redundant information for object recognition in computer vision applications. The sparse representation of the object's contour is a better choice for efficient and accurate recognition [14-15]. Intuitively, the sparse representation of the object's contour should be composed of the contour features. The scale invariance feature transform (SIFT) [16] has been used widely for feature extraction. However, it could only extract features based on the two-dimensional information, such as texture and color of the image instead of the one-dimensional contour of the object. In this letter, we explore a new direction of feature detection by slope difference distribution (SDD). We use the detected SDD features to form the sparse representation of the object contour and then use the SDD sparse representation to design a reference model for each object. At last, model matching is used for on line object recognition. Experimental results showed that the proposed method could achieve state of the art accuracy in several public datasets.

## II. SDD Feature Detection

The SDD features are defined as the points on the one-dimensional contour with greatest local slope differences. The SDD features include the SDD peak features that correspond to the clustering centers and the SDD valley features that correspond to the partitioning points. To detect the SDD features, we need to calculate the one-dimensional (1D) contour from the two-dimensional (2D) shape of the object. The centroid of the object is computed and the boundary of the object is extracted as the 2D contour.. Then, the 1D contour of the object $C_j^{1D}, j = 1,...,L$ is computed as the distance distribution from the centroid of the object to each point on the 2D contour as follows.

$$C_j^{1D} = \left[ \left(x_j - x_c\right)^2 + \left(y_j - y_c\right)^2 \right]^{1/2}, j = 1,...,L \quad (1)$$

where $L$ is the total number of points on the 1D contour.

Fig. 1 demonstrates the process of calculating the 1D contour from the 2D shape with a binary hand. After 1D contour is computed, it is normalized to be scale-invariant. Then, the normalized 1D contour $C_j^{1D}, j = 1,...,L$ is decomposed into different frequency components by discrete Fourier transform (DFT). The low frequency components constitute of the smoothed 1D contour and the high frequency components are the noise contained in the 1D contour. Therefore, only the low frequency part is kept and the high frequency part is removed. The 1D contour $C_j^{1D}, j = 1,...,L$ is transformed into the frequency domain by discrete Fourier transform (DFT).

$$F(k) = \sum_{j=1}^{L} C_j^{1D} e^{-i\frac{2\pi k j}{L}}; k = 1,...,L \quad (2)$$

The high frequency components in the Fourier spectrum of the 1D contour are removed by the following equation:

ZhenZhou Wang is now with the College of electric and electronic engineering, Shandong University of Technology, Zibo, Shandong 255000 China (e-mail: wangzz@sdut.edu.cn).

$$F'(k) = \begin{cases} F(k); k \leq W \\ 0; k > W \end{cases} \quad (3)$$

where $W$ is the cut-off frequency of the low pass DFT filter. After high frequency component elimination, the 1D contour is transformed back into spatial domain as follows.

$$C_j^{1D'} = \frac{1}{L}\sum_{k=1}^{L} F'(k) e^{\frac{i2\pi kj}{L}}; j = 1,\ldots, L \quad (4)$$

where $C_j^{1D'}$ is the smoothed 1D contour and its SDD is computed as follow.

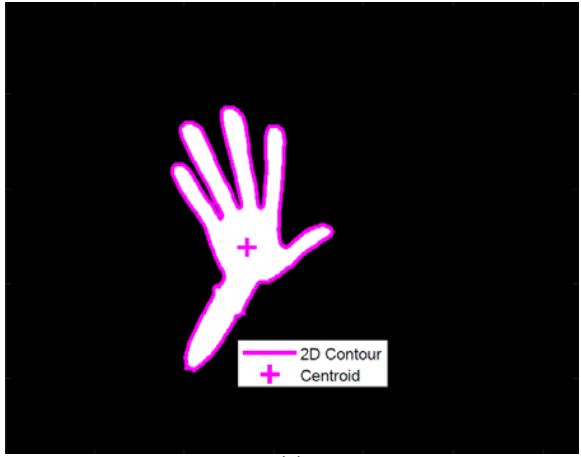

(a)

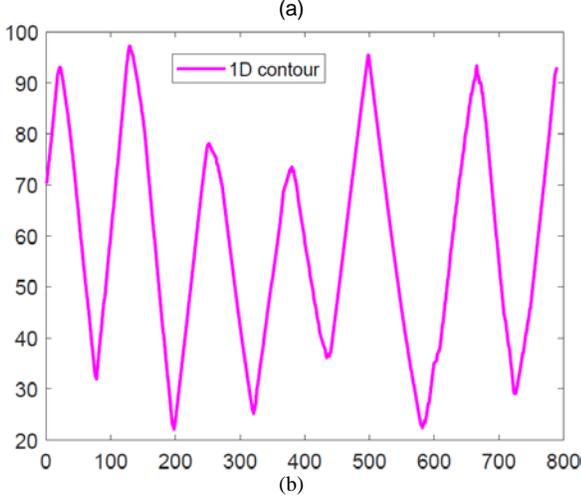

(b)

Fig. 1. Demonstration of calculating 1D contour from the 2D contour of the object with a binary hand (a), the extracted 2D contour from the binary hand; (b) the 1D contour calculated from the 2D contour.

Firstly, $N$ points $(x, C_x^{1D'}); x = j, j-1,\ldots, j-N+1$ on the left side of the point $(j, C_j^{1D'})$ and $N$ points $(x, C_x^{1D'}); x = j, j+1,\ldots, j+N-1$ on the right side of the point $(j, C_j^{1D'})$ are selected to fit two lines by the following equation.

$$y = aj + b \quad (5)$$

$a$ is the slope of the line and $b$ is a constant coefficient. $[a, b]^T$ is computed as:

$$[a, b]^T = (B^T B)^{-1} B^T Y \quad (6)$$

$$B = \begin{bmatrix} j+1-N & 1 \\ j+2-N & 1 \\ \vdots & \vdots \\ j-1 & 1 \\ j & 1 \end{bmatrix} \text{ or } \begin{bmatrix} j & 1 \\ j+1 & 1 \\ \vdots & \vdots \\ j-2+N & 1 \\ j-1+N & 1 \end{bmatrix} \quad (7)$$

$$Y = \begin{bmatrix} C_{j+1-N}^{1D'}, C_{j+2-N}^{1D'},\ldots, C_j^{1D'} \end{bmatrix}^T$$
$$\text{or } \begin{bmatrix} C_j^{1D'}, C_{j+1}^{1D'},\ldots, C_{j-1+N}^{1D'} \end{bmatrix}^T \quad (8)$$

Two slopes, $a_j^r$ and $a_j^l$ at the $j$th point $(j, C_j^{1D'}); j = N+1, N+2,\ldots, L-N$ could be obtained from Eq. (5). The slope difference distribution $s_j$ is computed as:

$$s_j = a_j^r - a_j^l \quad j = 1+N,\ldots, L-N \quad (9)$$

Setting the derivative of $s_j$ to zero and solve it, we get the positions of the valleys $V_i; i = 1, 2,\ldots, N_V$ with greatest local variations on the slope difference distribution and their magnitudes $M_i^V; i = 1, 2,\ldots, N_V$. We also get the positions of the peaks $P_i; i = 1, 2,\ldots, N_P$ and their magnitudes $M_i^P; i = 1, 2,\ldots, N_P$.

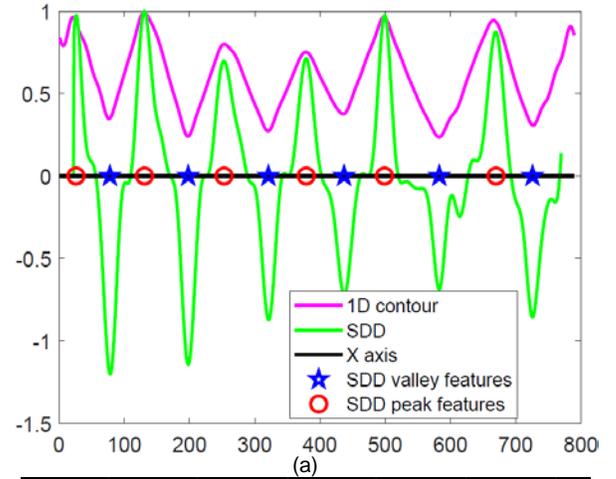

(a)

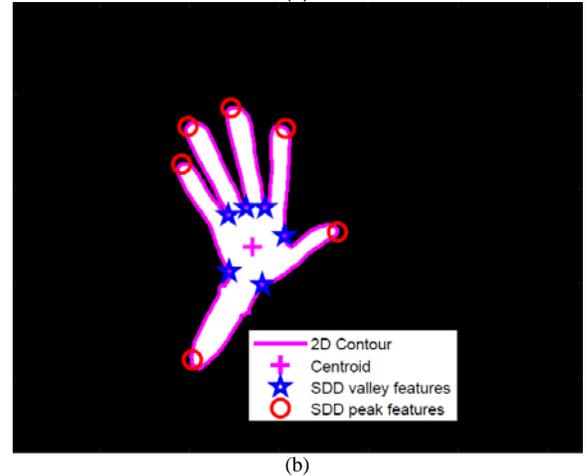

(b)

Fig. 2. Demonstration of calculating 1D contour from the 2D contour of the object with a binary hand (a), the extracted 2D contour from the binary hand; (b) the 1D contour calculated from the 2D contour.

The 2D SDD features on the 2D contour are computed based on the positions of the 1D SDD features on the 1D contour as follows. The 2D peak features are computed as:

$$P_i^{PF} = \{(x,y) \mid x = x_{F_i^P}, y = y_{F_i^P}, i = 1,2,...,N_F\} \quad (10)$$

The 2D valley features are computed as:

$$P_i^{VF} = \{(x,y) \mid x = x_{F_i^V}, y = y_{F_i^V}, i = 1,2,...,N_F - 1\} \quad (11)$$

Fig. 2 (a) demonstrates the process of calculating the SDD features from 1D contour and Fig. 2 (b) demonstrates the computed 2D SDD features by Eq. (10) and Eq. (11).

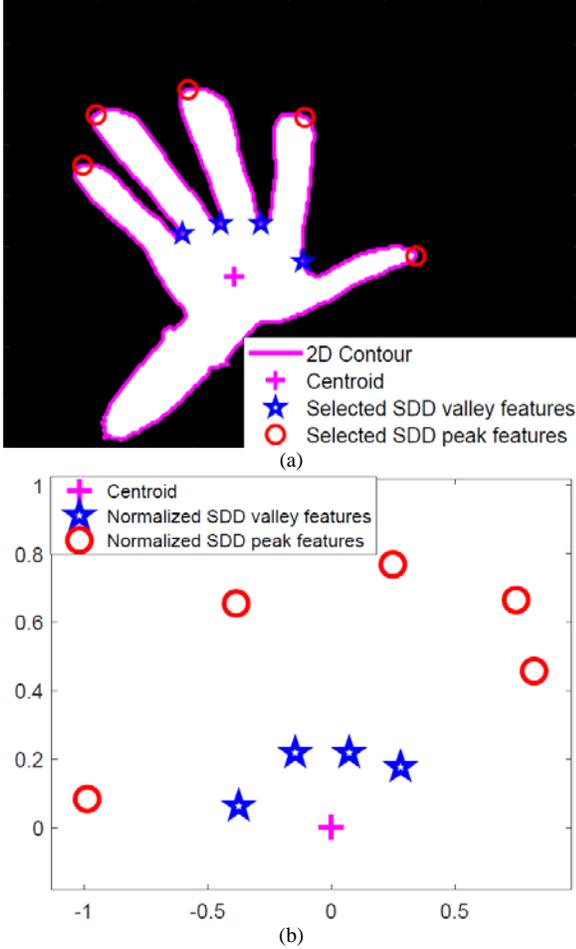

Fig. 3. Demonstration of the normalized SDD features (a), the selected 2D SDD features; (b) the normalized 2D SDD features.

## III. OBJECT RECOGNITION WITH SDD FEATURES

After the 2D SDD features are computed, specific feature selection rules are then designed to select the 2D SDD features based on the characteristics of the object. Then, the selected SDD features are normalized as follows. The selected 2D SDD peak features are normalized as:

$$P_i^P = (x_i^{PF} - x_c, y_i^{PF} - y_c), i = 1,2,...,N_F \quad (12)$$

$$P_i^{NP} = P_i^P / \max([P_1^P, P_2^P, ..., P_{N_F}^P]), i = 1,2,...,N_F \quad (13)$$

The selected 2D SDD valley features are normalized as:

$$P_i^V = (x_i^{VF} - x_c, y_i^{VF} - y_c), i = 1,2,...,N_F - 1 \quad (14)$$

$$P_i^{NV} = P_i^V / \max([P_1^V, P_2^V, ..., P_{N_F-1}^V]), i = 1,2,...,N_F - 1 \quad (15)$$

Fig. 3 demonstrates the selected and normalized SDD features. As can be seen, the normalized SDD features make up the sparse representation of the object. The reference model of each object is then built based on the normalized 2D SDD features.

During model matching, the minimum Euclidean distances between the automatically detected SDD features and all the reference models of the objects are computed as:

$$d_k = \arg\min_{\theta \in [0,45^o]} (d_\theta^P + d_\theta^V), k = 1,...,K \quad (16)$$

where $K$ is the total number of models in the dataset. $d_\theta^P$ is the average distance between the peak features in detected SDD contour sparse representation after rotating $\theta$ angle and the corresponding peak features in the reference model. $d_\theta^V$ is the average distance between the valley features in detected SDD contour sparse representation after rotating $\theta$ angle and the corresponding valley features in the reference pattern. $d_\theta^P$ and $d_\theta^V$ are computed as follows.

$$d_\theta^P = \frac{1}{N_F} \sum_{i=1}^{N_F} \left[ \left(x_i^{NP} - x_i^{RP}(k)\right)^2 + \left(y_i^{NP} - y_i^{RP}(k)\right)^2 \right]^{1/2} \quad (17)$$

$$d_\theta^V = \frac{1}{N_F - 1} \sum_{i=1}^{N_F - 1} \left[ \left(x_i^{NV} - x_i^{RV}(k)\right)^2 + \left(y_i^{NV} - y_i^{RV}(k)\right)^2 \right]^{1/2} \quad (18)$$

where $(x_i^{NP}, y_i^{NP}), i = 1,2,...N_F$ is the normalized coordinates of the automatically detected SDD peak features and $(x_i^{RP}(k), y_i^{RP}(k)), i = 1,2,...N_F(k)$ is the normalized coordinates of the SDD peak features in the $k$th reference model. $(x_i^{NV}, y_i^{NV}), i = 1,2,...N_F - 1$ is the normalized coordinates of the automatically detected SDD valley features and $(x_i^{RV}(k), y_i^{RV}(k)), i = 1,2,...N_F(k) - 1$ is the normalized coordinates of the SDD valley features in the $k$th reference model. The object is classified as the $k$th model with the minimum distance $d_k$ to it.

## IV. EXPERIMENTAL RESULTS

The quantitative classification accuracy of the proposed SDD contour sparse representation method on the NUS subset A is compared with state of the art method in Table 1. The proposed SDD contour sparse representation method achieved 100% recognition accuracy for all the tested images, which is significantly better than state of the art method.

Table 1. Quantitative comparison of the achieved recognition accuracies by the proposed method and state of the art method on the NUS subset A [17].

| Method | Accuracy (%) |
|---|---|
| [17] | 94.36 |
| **Proposed** | **100** |

The quantitative classification accuracy of the proposed SDD contour sparse representation method on the NI dataset is compared with state of the art method in Table 2. The proposed SDD contour sparse representation method achieved 100% recognition accuracy for all the images.

Table 2. Quantitative comparison of the achieved recognition accuracies by the proposed method and state of the art method on the NI dataset [18]

| Method | Accuracy (%) |
|---|---|
| [18] | 99 |
| **Proposed** | **100** |

The quantitative classification accuracy of the proposed SDD contour sparse representation method on the Kimia99 dataset is compared with state of the art method in Table 3. The proposed SDD contour sparse representation method achieved 100% recognition accuracy for all the images.

Table 3. Quantitative comparison of the achieved recognition accuracies by the proposed method and state of the art method on the Imia99 dataset [19].

| Method | Accuracy (%) |
|---|---|
| [19] | 97.17 |
| **Proposed** | **100** |

## CONCLUSION

In this letter, slope difference distribution is used to detect the features of the object's contour. To make the SDD computation feasible, the 2D object's contour is transformed to 1D contour by centroid-to-boundary distance calculation. A reference model is built based on the normalized 2D SDD features for each class of object. A SDD contour sparse representation method is proposed to recognize the object based on model matching. Experimental results with the open accessible datasets indicate that SDD has the potential for object recognition applications in the real world.